\documentclass{article}

\usepackage{microtype}
\usepackage{graphicx}
\usepackage{subfigure}
\usepackage{booktabs} % for professional tables

\usepackage[hyphens]{url}
\usepackage{hyperref}

\usepackage[accepted]{icml2024}

\usepackage{amsmath}
\usepackage{amssymb}
\usepackage{mathtools}
\usepackage{amsthm}
\usepackage{paralist}

\usepackage[capitalize,noabbrev]{cleveref}

\theoremstyle{plain}

\theoremstyle{definition}

\theoremstyle{remark}

\usepackage[textsize=tiny]{todonotes}

\icmltitlerunning{Position: A Call to Action for a Human-Centered AutoML Paradigm}

\begin{document}

\twocolumn[
%\icmltitle{Position Paper: A Human-centered Paradigm for Automated Machine Learning}
%\icmltitle{Position Paper: Rethinking AutoML Priorities}
\icmltitle{Position: A Call to Action for a Human-Centered AutoML Paradigm}



\icmlsetsymbol{equal}{*}

\begin{icmlauthorlist}
\icmlauthor{Marius Lindauer}{equal,hannover,l3s}
\icmlauthor{Florian Karl}{equal,iis,lmu,mcml}
\icmlauthor{Anne Klier}{iis}
\icmlauthor{Julia Moosbauer}{lmu,mcml}
\icmlauthor{Alexander Tornede}{hannover}
\icmlauthor{Andreas Mueller}{microsoft}
\icmlauthor{Frank Hutter}{freiburg}
\icmlauthor{Matthias Feurer}{lmu,mcml}
\icmlauthor{Bernd Bischl}{lmu,iis,mcml}
\end{icmlauthorlist}

\icmlaffiliation{lmu}{Ludwig-Maximilians-Universität München, Munich, Germany}
\icmlaffiliation{hannover}{Institute of Artificial Intelligence (LUH$\mid$AI), Leibniz University Hannover, Germany}
\icmlaffiliation{l3s}{L3S Research Center, Hannover, Germany}
\icmlaffiliation{iis}{Fraunhofer Institute for Integrated Circuits IIS, Fraunhofer IIS, Nuremberg, Germany}
\icmlaffiliation{mcml}{Munich Center for Machine Learning, Munich, Germany}
\icmlaffiliation{freiburg}{Albert-Ludwigs-Universität Freiburg, Freiburg, Germany}
\icmlaffiliation{microsoft}{Microsoft, Redmond, USA}

\icmlcorrespondingauthor{Marius Lindauer}{m.lindauer@ai.uni-hannover.de}
\icmlcorrespondingauthor{Florian karl}{florian.karl@iis.fraunhofer.de}

\icmlkeywords{Automated Machine Learning, Human-in-the-Loop}

\vskip 0.3in
]



\printAffiliationsAndNotice{\icmlEqualContribution} % otherwise use the standard text.

\begin{abstract}
Automated machine learning (AutoML) was formed around the fundamental objectives of automatically and efficiently configuring machine learning (ML) workflows, aiding the research of new ML algorithms, and contributing to the democratization of ML by making it accessible to a broader audience. 
Over the past decade, commendable achievements in AutoML have primarily focused on optimizing predictive performance.
This focused progress, while substantial, raises questions about how well AutoML has met its broader, original goals. 
In this position paper, we argue that a key to unlocking AutoML's full potential lies in addressing the currently underexplored aspect of user interaction with AutoML systems, including their diverse roles, expectations, and expertise. 
We envision a more human-centered approach in future AutoML research, promoting the collaborative design of ML systems that tightly integrates the complementary strengths of human expertise and AutoML methodologies. 
\end{abstract}

\section{Introduction}
\label{sec:introduction}
Over the last decade, Automated machine learning~\citep[AutoML, see][]{hutter-book19a,bergstra-jmlr12a,snoek-nips12a,thornton-kdd13a,escalante-ncs21a} has proven its potential to improve machine learning (ML) systems by automating parts of the data science workflow, in particular the selection and configuration of pipelines of ML algorithms of various sorts, by providing new and efficient hyperparameter optimization (HPO) procedures~\cite{feurer-automlbook19a,bischl-dmkd23a}, neural architecture search (NAS) methods~\cite{elsken-jmlr19a,white-arxiv23a} and the construction of powerful ensembles~\cite{erickson-arxiv20a}. 
AutoML success stories are numerous -- to name a few of them: the substantial contribution of AutoML to AlphaGo~\citep{chen-arxiv18a}; many AutoML systems with a total of over 100.000 downloads each month, e.g., AutoGluon~\cite{erickson-arxiv20a}, Auto-Sklearn~\cite{feurer-jmlr22a}, Auto-Weka~\cite{thornton-kdd13a}, Auto-Prognosis~\cite{alaa-icml18a}, SMAC~\cite{hutter-lion11a,lindauer-jmlr22a}; the routine usage of hardware-aware NAS for automatic design of neural architectures with hardware constraints in industry~\cite{benmeziane-ijcai21a}; learned optimizers like LION~\citep{chen-neurips23a}; the learned Swish activation function~\citep{ramachandran-iclr18a}; learned data augmentation strategies~\citep{cubuk-cvpr19a}; and prior-fitted networks (PFNs) for learning classification algorithms~\cite{hollmann-iclr23a}.
Because of that, AutoML research has grown rapidly over the last years, probably most evident 
in NAS~\cite{elsken-jmlr19a,white-arxiv23a}. 
At the same time, most big IT companies have developed large software packages enabling AutoML, including Google~\cite{golovin-kdd17a,song-automl22a}, Amazon~\cite{erickson-arxiv20a}, Meta~\cite{balandat-neurips20a}, IBM~\cite{wang-iciuic20a}, Oracle~\cite{yakovlev-vldb20a} and Microsoft~\cite{wang-mlsys21a}. 

Despite these successes, after more than a decade of research on AutoML, it is time to reflect on whether the AutoML community has achieved its original goals, whether those goals really addressed all the needs of all the targeted user groups in the first place and what is currently missing.
AutoML, in its current form, arguably aims at \begin{inparaenum}[(i)]
    % application
    \item accelerating the development of well-performing ML pipelines in applications by effectively addressing the problems of model selection and configuration (incl. neural architectures and hyperparameters);
    %research
    \item supporting research on new ML algorithms by automatically configuring the entire system and thus building the best possible system -- but also providing strong and appropriate baseline comparisons via essentially the same mechanism;
    % democratization
    \item contributing to the democratization of ML for domain experts with little to no ML expertise.
 \end{inparaenum}

Although many important challenges remain (e.g., regarding the scaling of AutoML to large foundation models, or the expressiveness of NAS methods), the potential and partial success of the first two goals was shown in many studies~\citep[and many more, see the references above]{chen-arxiv18a,ramachandran-iclr18a,guyon-automlbook19a,erickson-arxiv20a}, especially if we define ``performance'' narrowly in terms of ``efficiently optimizing predictive performance''. 

Nevertheless, one aspect that has not been sufficiently considered in large parts of AutoML research is that there are several user groups that could benefit from AutoML, each of which has very different needs and expectations: 
First, there are domain experts who would mainly like to communicate their general goals and domain knowledge to the AutoML system. 
They are typically very interested in understanding the final model -- or rather, a population-level understanding regarding their task that allows valid inferences about general relationships.
Then, there are ML practitioners and data scientists who deal with deeper and more technical and mathematical issues in applied model building. 
They usually like being in control of the ML process but want to automate away repetitive and mundane work (which human experts are less good at anyways), e.g., technical aspects of model selection (but not all of them) and especially HPO. 
Last but not least, there are ML researchers who focus on developing new ML approaches (and underlying theories). 
They nearly always need to be in full control and usually care much more about the effectiveness of the ML components; so explanations of HPO are more relevant to them than interpretations of the final models.
AutoML researchers or experts interact with an AutoML system mainly during its development and setup but could be considered an additional user group. They are mostly interested in information and visualizations to analyze the performance and behavior of AutoML systems.

However, several studies also showed~\cite{bouthillier-hal20a,hasebrock-arxiv23a,simon-cain23a,lee-ieeedeb20a} that AutoML has not fully permeated all these user groups 
and, thus, has not been able to reach its full potential. 
We attribute this to the following open challenges:

\begin{enumerate}
    \setlength\itemsep{0em}
    % too rigid and too inflexible
    \item Full AutoML systems were constructed too rigidly. Regarding their use by domain experts, automating the full process of ``data science'' is arguably a complex problem, and current AutoML systems provide an oversimplified and inflexible solution for this task, when it comes to e.g., the expression of (auxiliary) goals, model preferences and domain knowledge. 
    Likely, although more challenging, it would be more desirable for domain scientists to express such aspects via a natural interface into the system and also have results explained back to them in the same manner.  
    Furthermore, the design of AutoML systems as a software application (rather than a library) also complicates their use as a subcomponent in more complex systems and code bases, which is, in particular, relevant for data scientists and ML experts. 
    In such scenarios, HPO packages are far more convenient than monolithic AutoML software applications.
    \item Current AutoML systems address a narrow task in the data science process by mainly optimizing predictive performance. The shift from optimizing only the predictive model to optimizing full ML pipelines mitigated this problem to some degree, but data science encompasses a lot more than simply optimizing predictive performance. In many applications, aspects like interpretability, causality, fairness and robustness matter greatly, but these are hard to express in a single-objective metric a-priori. 
    Additionally, AutoML systems often cannot handle data organized in multiple tables or non-i.i.d. observations (time and curve data), which occur extremely often in practice. 
    \item AutoML is often not designed as an iterative process with human interaction but as a press-the-button-once system that returns a single design. 
    However, ML practitioners and data scientists are often unaware of hidden constraints and preferences a-priori that nonetheless matter for the task at hand.
    Often, this can only be figured out in an interactive process, where intermediate results are discussed with domain experts, implying that AutoML tools should support such an interactive workflow.
    While HPO tools already provide benefits to ML researchers, not all of their needs are fully addressed, especially the search for scientific insights.
    For example, many ML researchers require an understanding of hyperparameter sensitivity and the impact of new components. % and HPO of baselines to conduct solid empirical science~\citep{bouthillier-mlsys21a}.        
    \item Last but not least, AutoML tools would benefit from further efficiency improvements, especially for the challenging tasks at the cutting edge of ML research. When ML researchers are required to train the largest possible model on currently available hardware, they cannot afford many runs. At the same time, any kind of interactivity between users and AutoML would require a reasonably low response time from the AutoML tool.
\end{enumerate}

\textbf{Our Position:} We believe that most of these challenges are connected: they are caused by ignoring the interaction between users and AutoML systems, and the different roles, expectations, workflows, goals and valuable expertise of users. Although AutoML can support practitioners in many different ways, the strengths of AutoML approaches and users' expertise are complementary. \textbf{Therefore, we argue for a more human-centered paradigm in AutoML in this position paper, enabling efficient and collaborative design of ML systems by leveraging the best of both worlds, human experts, and systematic \mbox{AutoML}}. 

\section{Related Work}

To put our advocacy of a more human-centered AutoML paradigm into context, we give an overview of the ML community's shift towards a human-centered paradigm and the first few steps taken by the AutoML community. 

\subsection{Human-Centered Machine Learning}
\label{ssec:related_human_centered_ml}
In recent years, human-centered ML has gained significant momentum, driven by an increasing awareness of social and ethical implications of ML technologies. 
Leading the charge on human-centered ML are interpretable ML (including transparent decision making), interactive human-in-the-loop approaches and fair ML.
As the understanding of terms like interpretability, fairness and transparency is still evolving, these research fields develop and change fast.

Interpretable ML encompasses principles and methods that aim at offering explanations of why an ML model makes certain decisions~\cite{lipton-cacm18a,molnar-book22a}.
These include various approaches, like the development of interpretable models~\citep{rudin-nmi19a}, model-specific methods for deep neural networks~\cite{zhang-tetci21a}, model-agnostic techniques to visualize feature effects such as partial dependence plots~\citep{friedman-as01a} or accumulated local effects plots~\citep{apley-jrss20a}, methodologies for assessing feature importance~\citep{casalicchio-ecml19a, hooker-sc21a, ewald-arxiv24a}, and example-based explanations tailored to individual instances, such as Shapley values~\citep{lundberg-neurips17a,sundararajan-icml20a} or strategies like examining adversarial examples~\citep{goodfellow-iclr15a} and counterfactual explanations~\citep{wachter-hjlt18a, dandl-ppsn20a}, among others.
Given possible systematic errors or unwanted shortcuts taken by ML models, understanding them is crucial in high-stakes scenarios and legal requirements are increasingly mandating audits that rely on interpreting and verifying model behavior~\citep{eu2021aiact}.

Alongside transparency, fairness in ML has emerged as an important topic~\citep{barocas-book23a} to mitigate the risks of unlawful and socially detrimental discrimination.
In particular, it attempts to detect, avoid or at the very least mitigate biases of an ML model.
Finally, cooperative or interactive ML attempts to integrate human experts into the ML process. This is important from two perspectives: First, to increase performance by injection of expert knowledge and second, to increase the trust in these models by granting experts greater oversight of the systems and a deeper comprehension of the learning process~\cite{wu-fgcs22a}.

\subsection{Human-Centered AutoML}
As a human-centered paradigm increasingly permeates the field of ML, it becomes imperative to extend AutoML in this direction.\footnote{We note that human-centered can be understood in two ways: ``humans as users of ML'' and ``humans being impacted by ML''. Both are equally important, but from an AutoML perspective, we focus on the former view, while the latter is mostly out of scope for this work.}
While AutoML research mainly focuses on improving the computational efficiency of AutoML systems~\citep[c.f.][]{automl22, automl23}, 
there are already a few select papers proposing methods related to human-centered AutoML, especially in the realm of interpretable~\citep[see also Section~\ref{ssec:transparency_interpretability}]{biedenkapp-lion18a,ono-ieee20a,moosbauer-neurips21a} and interactive AutoML~\cite{anastacio-ppsn20a, souza-ecml21a, av-neurips22a, giovanelli-aaai24a, hvarfner_iclr24a}.
We highlight specific works alongside future research directions in Section~\ref{sec:open_problems}.
These works are complemented by publications from Human Computer Interaction (HCI) researchers, e.g. \citet{gil-iciui19a}, who collect interface requirements that need to be fulfilled to ensure that human-centered AutoML systems can recreate traditional ML workflows.
The position paper by \citet{pfisterer-arxiv19a} 
focuses on the consequences of the popularity of AutoML and possible interfaces for human-centered AutoML systems. 
Similar to this work, \citet{debie-cacm22a} argue that automated data science needs to be designed with humans in mind and should only support users, not replace them. However, they put more emphasis on the earlier (but arguably important and currently in AutoML neglected) stages of the data science workflow, i.e., data exploration and problem formalization concerns.
Finally, there are several user studies on AutoML, which give valuable insights into user requirements \cite{wang-pacmhci19a, drozdal-iui20a, crisan-chi21a, xin-chi21a, wang-arxiv21b, hasebrock-arxiv23a, sun-chi23a}. 
Some previous user studies focus on human-centered AutoML~\cite{lee-ieeedeb20a,khuat-fthci23a,xanthopoulos-edbticdt20a} and examine the current AutoML state-of-the-art and landscape from the user side, e.g., suggesting that AutoML systems be judged increasingly by how much users can interact with them.
Our work puts emphasis on investigating currently neglected research directions within the AutoML community that are related to the absence of a human-centered paradigm.

\section{The Case for Human-Centered AutoML}
\label{sec:position}

In our view, there are five main goals for AutoML tools:
\begin{inparaenum}[(i)]
    \item Predictive Performance,
    \item Optimization Speed,
    \item Transparency and Interpretability,
    \item Customizability and Flexibility, and
    \item Usability and Interaction.
\end{inparaenum}

The first two of these requirements (i.e., predictive performance and optimization speed) have been the primary focus of a large portion of AutoML research, and while further research is required (especially to scale AutoML for the age of foundation models), progress is already well underway for them based on multi-fidelity optimization~\citep{li-jmlr18a,falkner-icml18a,wistuba2022supervising,kadra2023scaling}, exploiting user priors~\citep{hvarfner-iclr22a,mallik-icml23a,hvarfner_iclr24a}, and transfer learning~\citep{wistuba-ml18a,feurer-arxiv18a,wistuba-iclr21a}. In contrast to a fully automated approach, we note that a human-centered approach can incur time costs for humans.  Both leveraging human expertise and automation via efficient optimization can aid in obtaining increased efficiency of machine learning workflows; we argue that combining them will not result in increased costs but, through their complementary nature, achieve this common goal. A more detailed discussion on this tradeoff can be found in Appendix Section~\ref{app:cost}.
With that in mind, this paper focuses on the last three goals, that are in our opinion currently understudied in the AutoML community.

Overall, the machine learning workflow CRISP-ML(Q), as described by \citet{studer-mdpi21a}, consists of six major phases: \begin{inparaenum}
    \item business and data understanding,
    \item data preparation,
    \item model engineering,
    \item model evaluation,
    \item model deployment, and
    \item model monitoring and maintenance.
\end{inparaenum} 
Figure \ref{fig:crisp} visualizes these phases and puts (interactive) AutoML research in the context of the CRISP-ML(Q) workflow.
Due to the inherent complexity, even data science experts need to iterate this workflow, potentially returning to a much earlier stage if the problem has been understood better \cite{xin-arxiv18a}.
Rapid prototyping, involving swift development and model testing, helps establish an initial baseline. Subsequently, users engage in an often lengthy, iterative trial-and-error process where they experiment with different configurations of the ML workflow to achieve satisfactory outcomes. 
Therefore, it is not surprising that AutoML is often used for establishing first baselines and model refinement after further insights about the problems at hand are obtained. 
This iterative and collaborative nature has been overlooked when designing past AutoML systems, and incorporating it explicitly into future systems promises to increase user productivity. 

We posit several hypotheses about key insights that have not yet been sufficiently addressed in AutoML.

\subsection{Hypothesis 1: Transparency and Interpretability Are Key for ML and AutoML in Many Applications and on Many Levels}
\label{ssec:transparency_interpretability}

\begin{figure}[tb]
\centering
  \includegraphics[width=0.99\columnwidth]{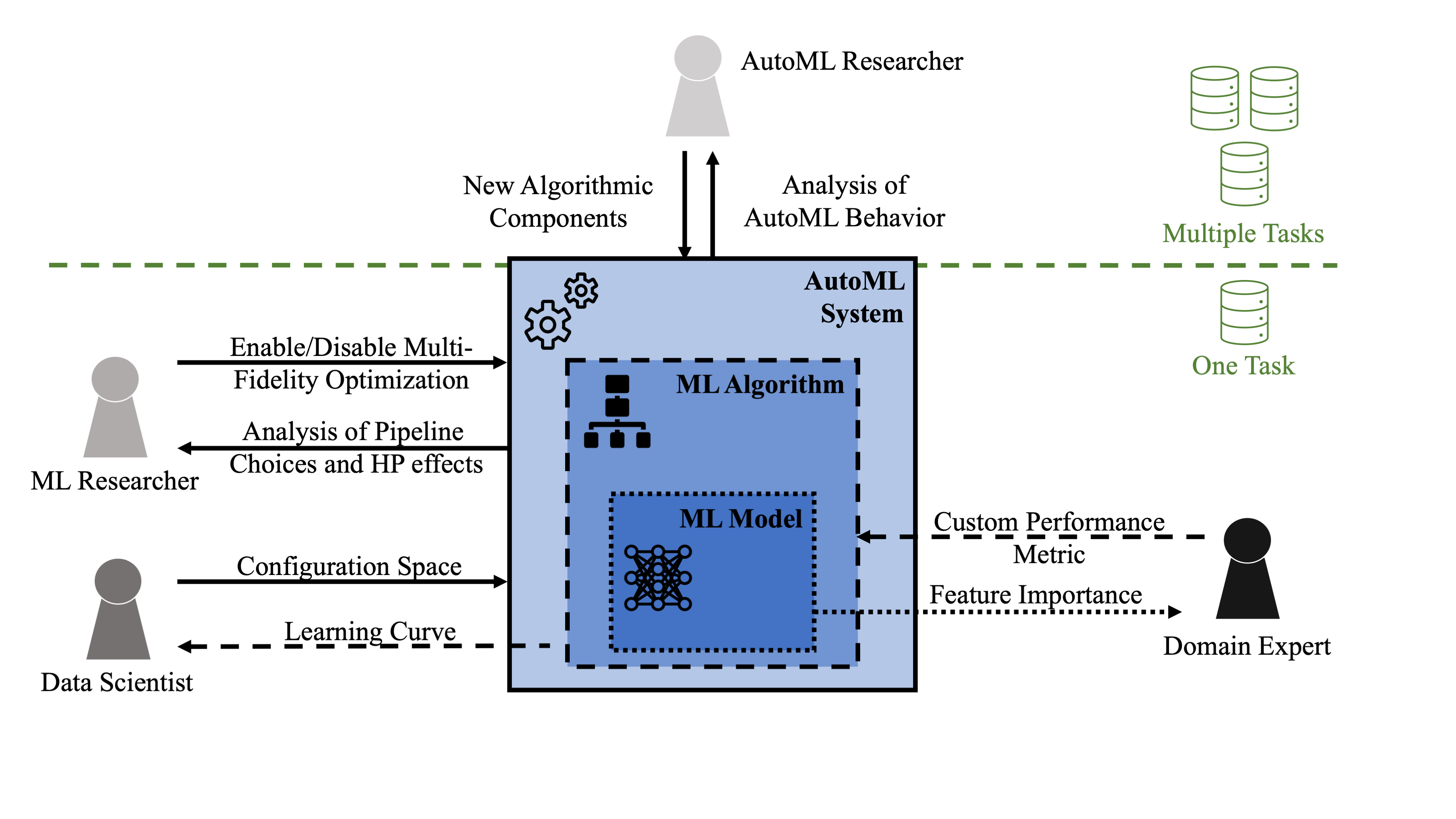}
  \caption{Selected (one-time) interactions of different user groups with AutoML, ML algorithms and models.}
  \label{fig:levels_of_information}
\end{figure}

Transparency and interpretability are closely linked and are an important source of trust for users in AutoML systems~\citep{wang-pacmhci19a,drozdal-iui20a}. 
Some studies report interpretability as a concrete requirement requested by users \citep{xin-chi21a,hasebrock-arxiv23a,sun-chi23a, wang-pacmhci19a}.
E.g., a lack of interpretability of AutoML tools led to study participants choosing manual development for projects with higher stakes~\citet{xin-chi21a}. 
We expand the three levels of interpretability of \citet{moosbauer-diss23a} by a fourth level, see Figure~\ref{fig:levels_of_information}. 

\textbf{1 - ML model interpretability} deals with the interpretability of the final model and addresses common questions within the field of interpretable ML. 
Often, understanding the underlying relations in the data-generating process is the primary target of modeling, instead of simply creating a black box predictor~\citep{shmueli-statsci10,bzdok-natmet17}, as, e.g., stated by data scientists participating in the study by \citet{wang-pacmhci19a}.
Even if maximizing predictive performance is the primary objective, being able to explain and audit a model is usually of high value.\\
\textbf{2 - ML algorithm interpretability} focuses on the behavior of the learning algorithm, aiming to explain aspects like convergence and the interpretation of learning curves.\\
\textbf{3 - AutoML system interpretability} deals with understanding why and how an AutoML system has chosen a certain element or pipeline and how hyperparameters influence the final result.
This question is usually relevant to a per task / per data set context.\\
\textbf{4 - AutoML comparative performance interpretability} deals with the performance of the AutoML system itself and could help users understand e.g., what algorithmic choices on the AutoML level may improve an AutoML system for certain applications and why~\cite {dang-gecco17a,lindauer-dso19a,moosbauer-telo22a}. 
These questions naturally arise in scientific contexts, where performance comparisons over multiple data sets are of interest.

The relevance of the above requirements varies among user groups. For example, domain experts may prioritize transparency and interpretability of the model returned by the AutoML system, whereas  ML practitioners are also interested in understanding why a certain model was returned. 

In contrast, ML researchers are often more interested in ablation studies, the robustness of model performance concerning hyperparameters and the effect of hyperparameters. Since AutoML typically collects a lot of data about the performance of different configurations, this data can build the foundation to create such insights~\cite{hutter-icml14a,fawcett-heu16a,biedenkapp-aaai17a,biedenkapp-lion18a,moosbauer-neurips21a,sass-realml22a,watanabe-ijcai23a,theodorakopoulos-arxiv24a}. 
At the same time, traditional interpretable ML methods might not be directly applied to performance data collected by AutoML systems because the AutoML process is typically biased towards well-performing configurations and thus, the performance data is biased too~\cite{moosbauer-neurips21a,segel-automl23a}, at least when one is interested in a global analysis across the whole configuration space.

\subsection{Hypothesis 2: Customizability and Flexibility Are Essential to Leverage the Potential of AutoML for Different User Groups}
Different AutoML tools operate on different levels of abstraction, each providing varying degrees of customization. Fully automated AutoML systems usually provide the least flexibility but the largest amount of automation. 
NAS approaches generally allow users to provide a pipeline specification but are still limited in their expressiveness and the building blocks they provide. HPO approaches provide the most flexibility but require users to set up pipelines, configuration spaces and evaluation themselves. 
However, even for fully automated systems, users might have to configure them correctly for their particular use case to obtain the best performance possible~\cite{lindauer-dso19a,moosbauer-ieee22a,feurer-jmlr22a,neutatz-vldbj23a}.

Recent studies showed that the ML practitioner user group criticized current AutoML tools in terms of these customization options~\citep{xin-chi21a,sun-chi23a,zöller-acm23a}. In the study by \citet{xin-chi21a} where users rated attributes of current AutoML tools, customizability was among the qualities that received the lowest ratings.
Studies however identified different needs when it comes to different customization options. 
Both \citet{hasebrock-arxiv23a} and \citet{wang-pacmhci19a} found that current tools for HPO only optimize for predictive performance, whereas practitioners have several additional objectives, such as increasing model comprehension, which could be accounted for using multi-objective optimization~\citep{karl-evolearn23a, horn-ssci16a, binder-gecco20a}. 
Furthermore, \citet{zöller-acm23a} found that users have a concrete need to adapt the configuration space to include what they learned from previous AutoML runs, which also very directly hints at the fact that users have a strong need for more interactive workflows, as requirement specifications in data science projects are usually not perfectly precise in a first try.
A partial reason for this request is likely also the desire to inject domain knowledge and/or to speed up the optimization, for example, through warmstarting~\cite{anastacio-ppsn20a,souza-ecml21a,hvarfner-iclr22a,mallik-icml23a,hvarfner_iclr24a}. 

Overall, the users of AutoML are diverse, and different applications demand different functionalities of AutoML and, specifically, different types of user interactions. For example, a biologist's primary objective might be accurate predictions of certain processes for which they would like to contribute domain expert knowledge (e.g., a specific kernel or distance function for gene data), while a bank employee, in addition, has to satisfy regulations on interpretability. In the case of a data scientist conducting unsupervised learning, an interactive AutoML approach based on preferences might be required if the ideal performance metric cannot be defined explicitly. For a more in-depth description of the aforementioned applications we refer the interested reader to Appendix Section~\ref{app:users}.

Due to this diversity in users and applications, another possible approach to AutoML, aside from a customizable platform solution, is a modular one. Frameworks like \textit{GAMA}~\cite{gijsbers-ecml21} or the recently proposed \textit{AutoML Toolkit}\footnote{\url{https://github.com/automl/amltk}} aim to provide a toolbox that allows users to design the AutoML solution geared towards their individual applications and beyond simply choosing one black-box optimization algorithm over another. This design philosophy is related to human-centered AutoML in two ways. First, a similar design philosophy could also be extended to interactive, explainable and overall human-centered AutoML by including modules that correspond to certain interactions (position in the data science lifecycle, interpretability level etc.); to the best of our knowledge, a framework adopting this approach specifically for human-centered elements of AutoML has not been proposed. Second, simply by adopting a modular approach, the AutoML process itself becomes more human-centered. Conscious decisions have to be made about design choices of the AutoML solution, which cannot or should not be automated, such as fairness~\cite{weerts-jair24}. At the end of the day, this will lead to a new abstraction layer of how to build ML systems, hiding many tedious and error-prone design decisions such as hyperparameters and allowing users to focus on the essential decisions by combining different high-level modules, leading to responsible and trustworthy use of (Auto)ML.

\subsection{Hypothesis 3: AutoML Tools Have to Integrate with the Data Science Workflow Allowing for an Iterative Interaction with the User}
The user experience, i.e. usability, and the options for interacting with current AutoML tools are important and, as for some of the other hypotheses, corresponding requirements vary among user groups \citep{xin-chi21a,wang-arxiv21b,crisan-chi21a}.
\citet{hasebrock-arxiv23a} observe that an important reason for ML practitioners choosing grid search or random search over Bayesian optimization for HPO is the ability to easily integrate these methods into their workflow.
We speculate that this is due to additional work that is required to integrate more advanced HPO, especially iterative and synchronous techniques, such as Bayesian optimization, into the technical workflow. 

Importantly, the degree of automation or the number of potential interaction points, respectively, for a user drastically depends on the user group \citep{lee-ieeedeb20a,crisan-chi21a,wang-arxiv21b}. 
For example, \citet{crisan-chi21a} show that users with a higher technical expertise tend to prefer less automation than users with a less technical background. Similarly, \citet{wang-arxiv21b} highlight that the desired level of automation over the different stages of the machine learning workflow varies for different data science-related roles.

Interaction with an AutoML system can typically happen on the levels introduced in Section~\ref{ssec:transparency_interpretability} and can take various forms: a user may desire to give input to, receive output from, or mutually interact with the AutoML system throughout the optimization process, and the details of the requirements for interaction depend again on the user group. 
For example, domain experts may want to ingest their expert knowledge to inform the ML algorithm, e.g., we might need to learn the expert's internal loss, which they cannot precisely specify, or to restrict a Pareto set to guide the optimization process towards practically desired directions. 
Another example would be users including their preferences to guide the optimization process either implicitly through prior beliefs on promising pipelines~\cite{mallik-icml23a,hvarfner_iclr24a} or explicitly by relative preferences between proposed pipelines~\cite{kulbach-ecai20a,giovanelli-aaai24a}.
The latter was, e.g., recently integrated in \textit{Optuna}~\cite{akiba2019optuna} since Version 3.4 through preferential Bayesian optimization.

We refer the interested reader to Appendix Section~\ref{app:mlops} for a discussion of how machine learning operations (MLOps) relates to human-centered AutoML.

\subsection{Hypothesis 4: Since Human Experts Are Essential to ML Processes, AutoML Will Only Reach Its Full Potential by Collaborating with Them}
\label{ssec:importance_of_human_experts}

Experts - mostly in the form of domain experts and data scientists - shape the lifecycle of an ML model in diverse ways.
\citet{khuat-fthci23a} even argue: ``systems cannot be considered optimal if they do not welcome and make use of optional human input''.
The time of these experts is, however, a precious resource in the development of ML solutions; human experts have to be integrated into the AutoML process, but the effort on their part should be minimal. We base the discussion on the groups identified in Section~\ref{sec:introduction}.

\textbf{Domain experts} can provide invaluable context through their knowledge of the application domain. 
The user study by \citet{xin-chi21a} concludes that participants mainly use their domain knowledge for data pre-processing steps, such as feature engineering, and for validating the resulting model. 
\citet{khuat-fthci23a} detail many possible uses of domain knowledge in every step of an ML workflow, ranging from defining success criteria in the beginning to selecting a configuration space for model search and monitoring the deployed model for possible biases.
It can easily be argued that the more domain knowledge a practitioner has, the greater their need to customize and influence the AutoML system~\cite{wang-chi19a,sun-chi23a}.

A lot of data science and ML projects are successful precisely because of the collaboration of domain experts and data scientists~\cite{mao-acmhci19a}.
So, even if data scientists can work more effectively with AutoML tools, domain experts are still integral to the success of ML projects.
A common workaround to handle the perceived inflexibility of AutoML systems is often
to inject domain knowledge in the optimized objective of the system in a rather technical manner~\cite{sun-chi23a}.  
Based on this, the authors suggest that instead AutoML systems should be developed either for specific domains and applications and/or support an interactive approach so that users can supply their domain expertise into the process.

\textbf{ML experts / data scientists} are largely receptive to the advantages of using AutoML in a supportive role as outlined in Section~\ref{ssec:automl_dependency}.
In practice, ML projects are rarely represented by a linear workflow where a dataset is presented and a model chosen, but through an iterative process.
Often, new data is acquired and labeled because of the information gained through baseline performances, which then may require an adaption in model selection and tuning.
A data scientist may also decide to include unlabeled data, which requires special modeling techniques.
The complexity and variety of these processes make navigating a given ML project's optimal workflow challenging.
Many data scientists agree that these types of strategic decisions or the investigative mindset of an expert cannot be fully automated in the near future~\cite{wang-pacmhci19a,debie-cacm22a}.
By the same reasoning, if human experts do not fully comprehend the process leading up to finding and training an optimal model, a lot of information will be lost that might have sparked further improvements for future iterations.

\textbf{ML researchers} are the main factor for moving ML forward.
They have invaluable knowledge about their fields of expertise that would be foolish to ignore by any AutoML system.
While AutoML systems could, in principle, learn to self-evolve and outcompete human ML researchers~\citep{clune-arxiv19a,huang2023benchmarking}, we believe it to be unlikely that AutoML systems alone will yield major novel research breakthroughs (on the level of discovering transformers) without human experts being closely in the loop anytime soon.
Nevertheless, we fully expect that, based on the increasing computational efficiency of AutoML and its ability to seamlessly reason about thousands of short and cheap experiments (with downscaled models, less data and fewer epochs, combined with extrapolation models), it will become ever more standard for ML researchers to use AutoML in their research, leading to a speedup in ML progress by AutoML.
Early examples of this already exist in the literature, with the identification of 
new state-of-the-art deep neural network architecture variants~\cite{so-neurips21a}, activation functions~\cite{ramachandran-iclr18a}, variants of weight updates~\cite{real-icml20a} or neural optimizers~\cite{andrychowicz-neurips16a,chen-neurips23a}, and we will likely see more such works in the future as AutoML methods become increasingly powerful and convenient to use. 

\textbf{Across user groups}
In general, interaction and communication between the user and the AutoML system leads to increased trust~\cite{crisan-chi21a,drozdal-iui20a}, which makes a human-centered paradigm necessary for the widespread adoption of AutoML.
This is especially challenging for those without a background in ML \cite{wang-pacmhci19a}, but obviously, this depends on the form of communication used (with the current form of rather mathematical and model-based communication being harder to understand than potential natural language output).
Another key role that human experts fill in ML projects and that goes hand in hand with the issues of trust and transparency, is that of a regulatory body.
Sanity checks (e.g., ``maybe there is data leakage because this model performs too well''), ethical concerns (e.g., ``this model may be biased against a certain population subgroup'') and safety standards (e.g., trained physicians giving the final approval for a course of therapy suggested by ML) are all important to ensure the quality, fairness and safety of ML~\cite{khuat-fthci23a}. 
Moreover, for certain aspects such as fairness, it is highly debatable, whether they can be cast into a metric and then automatically optimized~\citep{weerts-jair24}, which further underscores the need for a human-in-the-loop for model validation. 

Taking all these factors into account, we believe that AutoML, in its current form, is bound to reach a lower ceiling than it could reach if a human-centered paradigm is adopted - in terms of performance, number of applications as well as ethical and safety standards. Thus, allowing interactions, such as providing expert priors on well-performing designs, updates of data or the design space will be crucial to bringing humans back into the AutoML loop.

\subsection{Hypothesis 5: Human-Centered AutoML Empowers Users Instead of Making Them Dependent on a System They Do Not Understand}
\label{ssec:automl_dependency}

Automation, particularly AutoML, can bear risks when used without care, e.g., discrimination of groups. This can be especially problematic for users of AutoML as evidenced by \citet{zöller-acm23a} who observe the trend of users with little ML knowledge overestimating their understanding of a model proposed by an AutoML system. 
This can be seen as an instance of automation bias, i.e., the tendency to place too much trust in automated recommendations~\cite{skitka-ichcs99a}. 
The combination of this overly high level of trust and lower barriers to using ML may lead to ML being used for more and more applications where it may not be desirable. 
This concern of automating bad decisions is shared by participants in the user study by \citet{crisan-chi21a}. 
Naturally, one would hope that manually developed ML applications (without the use of AutoML) would include better oversight of experts and thus allow for several stopping points if the ML application turns out to be problematic.
We further discuss the effects of bias of humans and automation in Appendix Section~\ref{app:bias}.

With a shift of the focus from pure automation to a human-centered approach, the next generation of AutoML tools can avert many of these dangers and potentially even lead to more positive changes. 
In particular, we believe that more research in the direction of human-centered AutoML has the potential to empower people instead of making them data-science-illiterate. 
With the original promise of AutoML of lowering the entry barrier of ML for users, transparency and interpretability allow users to understand which parts have been automated and why certain outputs are obtained. 
As such, human-centered AutoML could focus on (i) the automation of the tedious and error-prone repetitive task of choosing a well-performing model and optimizing the corresponding hyperparameters and (ii) the interpretability of this process and its outcome. 
It allows users to focus on those parts of the data science workflow where less automation is possible or desirable. 

In the same spirit, human-centered AutoML also allows different data science teaching paradigms: future data scientists might be able to focus much more on the data at the start of their education instead of manually trying tens of different learners without gaining valuable insights by doing so. With the emergence of data-centric artificial intelligence / machine learning (DCAI/DCML), which (re-)emphasizes the importance of data quality, this has even more merit.
Additionally, the interpretability of results combined with a good explanation interface could allow users to learn something about the modeling task from the output of an AutoML tool. 
As such, while they focus on other parts of the workflow, they can still benefit from applying the tool by acquiring knowledge. 
This vision is also prevalent in the recent literature, with \citet{wang-pacmhci19a} arguing that AutoML should also fulfill an educational role in the future and \citet{xin-chi21a}, who find that several users took models given by AutoML tools as an opportunity to learn more about ML techniques. 

\begin{figure*}[ht]
    \centering
    \includegraphics[width=1.5\columnwidth]{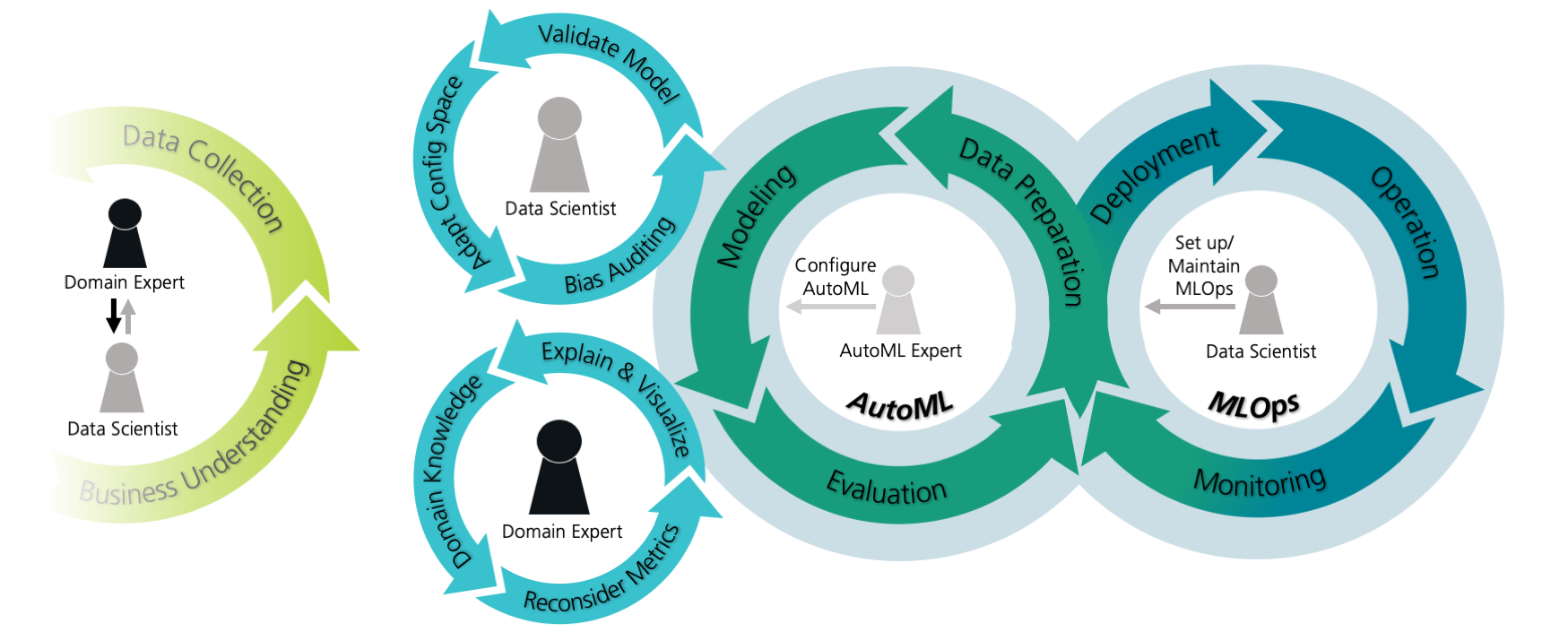}
    \caption{CRISP-human-centered AutoML Cycle, inspired by \cite{crisp}, with a focus on iterative interactions.}
    \label{fig:crisp}
\end{figure*}

\section{Ongoing Work and Future Directions}
\label{sec:open_problems}

In the following, we discuss existing and modern work in human-centered AutoML and highlight several specific opportunities for new research in this area.

\subsection{Existing Work on Trust and Interpretability in the Context of AutoML}
In Section~\ref{ssec:transparency_interpretability}, we argued that interpretability and transparency are some of the core features that facilitate trust in AutoML systems.
In fact, during the last few years, a growing amount of publications targeting interpretability in AutoML have been published, usually in the context of extending HPO.
On the first interpretability level, multi-objective optimization~\citep{karl-evolearn23a} can be used for finding tradeoffs between accurate and less complex/interpretable models~\citep{igel-emco05a,molnar-ecml19a,binder-gecco20a,schneider-gecco23a} or fairness-aware models~\citep{perrone-aies21a,weerts-jair24}.
On the third interpretability level, multiple approaches exist to increase hyperparameter interpretability. 
Approaches include measuring hyperparameter importance~\citep{hutter-icml14a,jin-arxiv22a, watanabe-ijcai23a}, or expressing hyperparameter effects \cite{moosbauer-neurips21a,segel-automl23a}. 
Furthermore, \citet{li2020-automl20} constrain the points explored by Bayesian optimization to ensure explainability.
Lastly, on the fourth level, there is also work on analyzing configuration spaces for HPO and hyperparameter importance across datasets~\citep{rijn-kdd18a,probst-jmlr19a}, but also work on understanding the optimization algorithms themselves~\citep{dang-gecco17a,lindauer-dso19a,moosbauer-telo22a}.

Similarly, multiple tools directly explain AutoML systems or their outputs (compared to the more methodological work above). 
Model LineUpper \cite{narkar-iciui21a} allows users to compare candidate models on multiple information levels. PipelineProfiler \cite{ono-ieee20a} visualizes ML pipelines produced by AutoML systems. DeepCAVE \cite{sass-realml22a} and XAutoML \cite{zöller-acm23a} can visualize optimizer runs.
ATMSeer~\cite{wang-chi19a} explains through visualization what models have been evaluated and how they performed; it also offers a visualization tool to support users in adapting the search space.

Nevertheless, we believe that there is still quite a bit of road ahead. 
In particular, we encourage the AutoML community to continue working towards understanding what trust, transparency, interpretability and related terms signify in the context of AutoML specifically, particularly when not applied to the model-level but rather the higher levels as discussed in Section \ref{ssec:transparency_interpretability}.

\subsection{Bridging the Gap Between Algorithmic and HCI AutoML Research}
Above, we focused on technical publications on methods for explainability and interaction in the context of AutoML.
At the same time, the field of HCI has produced insightful data on user needs and recommendations on how to fulfill them, both for human-centered AutoML systems specifically \cite{gil-iciui19a, khuat-fthci23a} and for human-AI interaction in general \cite{amershi-chi19a, yang-disc18a}, often with a focus on trust and interpretability \cite{liao-chi20a, hoffman-arxiv18a, vössing-isf22a}.
We encourage more collaboration between these two communities.
Few publications have tackled the issues from both sides, but some examples exist. In particular, \citet{zöller-acm23a} examine users' needs for visualization w.r.t. transparency in AutoML and then propose a framework that satisfies these requirements.
Although some survey participants felt overwhelmed by the amount of information presented in their framework XAutoML, it is certainly an important step in the right direction.

\subsection{Better Interfaces for AutoML}
Thus far, a major obstacle to harnessing the power of ML for people with a non-technical background has been the lack of a sufficiently intuitive way for users to formulate their tasks, domain knowledge, preferences or constraints.
In current systems, the interface language is largely mathematical/statistical regarding problem, goal and model specification, and program code in terms of implementation.

This constitutes an obstacle for domain experts, who, e.g., might have a good understanding of how ML models should be evaluated due to their domain knowledge, but may not be able to specify, e.g., the loss or performance metric as a precise formula. 
\citet{lee-ieeedeb20a} and \citet{bakshy-automl23} outline the additional problem that domain knowledge can be so complex that it is difficult to map it to simple constraints.
An interesting approach to tackle this issue might be preference learning or preferential optimization \cite{kulbach-ecai20a, diaz-ejor21a, ungredda-gecco23a, giovanelli-aaai24a}, where users are only asked to provide relative feedback by indicating their preference for one outcome over another.
This way, they can indirectly communicate important aspects of evaluation but do not have to formally specify objectives and constraints.

Arguably, large language models (LLMs) offer another opportunity to provide an easy-to-use interface to AutoML methods~\cite{tornede-arxiv23a}.
The success of LLMs with user groups beyond data science and ML suggests that such a text-based interface increases usability for a wide variety of potential users which could be a key piece of the puzzle for widespread adoption of AutoML systems.
LLMs may provide exactly what is needed to allow AutoML to provide the power of ML to many potential users without sacrificing the benefits of a human-centered paradigm: a suitable interface. This is reflected by \citet{karmaker-acmcsur21a}, who see a natural language interface as a prerequisite for domain experts to interact comfortably with AutoML tools.
First attempts have been made to use LLMs as an interface to AutoML systems for feature engineering~\cite{hollmann-neurips23a} or even full ML pipelines~\cite{zhang-arxiv23a}.
We encourage the community to explore this topic further and to facilitate an interactive approach to AutoML through LLMs and other generative multi-modal models. 
In this context, it could be very promising to couple very fast AutoML systems, such as TabPFN~\citep{hollmann-iclr23a}, with LLMs to allow for a highly interactive user experience.

\subsection{From Human-Centered AutoML to Human-Centered Automated Data Science}
As with AutoML research in general, most interactive and explainable AutoML approaches are focused on the modeling part of a machine learning project; 
but AutoML may well be extended beyond modeling. In fact, \citet{debie-cacm22a} have argued along similar lines as we do in the direction of a human-centered automated data science approach. They argue that some parts of the data science workflow rely on human input (e.g., domain expertise in data acquisition and labeling) and require human oversight (e.g., task definition through business understanding). 
In fact, data scientists often argue that the early stages of the data science workflow are the most important and time consuming ones; in contrast, the modeling part often contributes considerably less or is less time-consuming~\cite{press-forbes16a}.
Thus, there is great potential in taking a human-centered approach to AutoML across the data science lifecycle. 
An initial work in this direction is CAAFE~\citep{hollmann-neurips23a}, a human-understandable feature engineering approach for semi-automated data science.
Furthermore, an AutoML solution could help users navigate the often non-linear and iterative process through the data science lifecycle.
As shown in Figure~\ref{fig:crisp}, AutoML can support users in certain aspects regarding model development but requires a human-centered component to intervene if necessary.
A human-centered AutoML solution could furthermore help users decide if a data-centric paradigm is most promising moving forward (e.g., labeling of additional data) or a model-centric paradigm (e.g., allocating additional computing resources to modeling) and thus support users in decision making.

\section{Conclusion}
\label{sec:conclusion}
AutoML had a great success story over the last few years, with a plethora of impressive community achievements.
Nevertheless, there is still potential for improvement and further research, in particular in the underexplored area of human interaction with AutoML systems. 
In this position paper, we proposed and elaborated on a more human-centered approach to AutoML. 
In particular, we have formed a series of hypotheses on the need for more transparency, interpretability, customizability, flexibility, usability and the integration of human-centered aspects into AutoML research in general.
Based on these hypotheses, we analyzed the status quo and encouraged specific future research topics to move the field into the direction we anticipate with this work.
In particular, we encouraged more work in 
interpretability of AutoML, better user interfaces for AutoML, 
LLMs as an interface to AutoML and extending the human-centered automation paradigm to other parts of the data science workflow.

\newpage

\newpage % Acknowledgments can also optionally be included in another unnumbered section; both Impact Statement and Acknowledgments are excluded from the 9-page limit.

\section*{Impact Statement}

Since machine learning is arguably becoming extremely important in research, applications, and industry, and potentially even our personal lives, it is also crucial to push for a democratization of machine learning. 
AutoML has always aimed to contribute to this democratization, but we believe that the research on AutoML and its tools has underappreciated the requirements and expectations of its different user groups. 
We believe that this will generate new important research stimulus and eventually lead to a new generation of AutoML tools that are more useful to their users. 
Since human-centered AutoML tools will also enable to take ethical considerations into account (e.g., interpretability and transparency), we further believe that this paradigm will contribute to the responsible use of machine learning.

\section*{Acknowledgements}
We thank Giuseppe Casalicchio for valuable discussions and input on interpretable ML and connections to AutoML.

Alexander Tornede and Marius Lindauer acknowledge funding by the European Union (ERC, ``ixAutoML'', grant no.101041029). Views and opinions expressed are, however, those of the author(s) only and do not necessarily reflect those of the European Union or the European Research Council Executive Agency. Neither the European Union nor the granting authority can be held responsible for them. Furthermore, Marius Lindauer acknowledges support from the Federal Ministry of Education and Research (BMBF) under the project AI service center KISSKI (grantno.01IS22093C). Anne Klier and Florian Karl acknowledge support by the Bavarian Ministry of Economic Affairs, Regional Development and Energy through the Center for Analytics – Data – Applications (ADA-Center) within the framework of BAYERN DIGITAL II (20-3410-2-9-8).

\includegraphics[height=5cm]{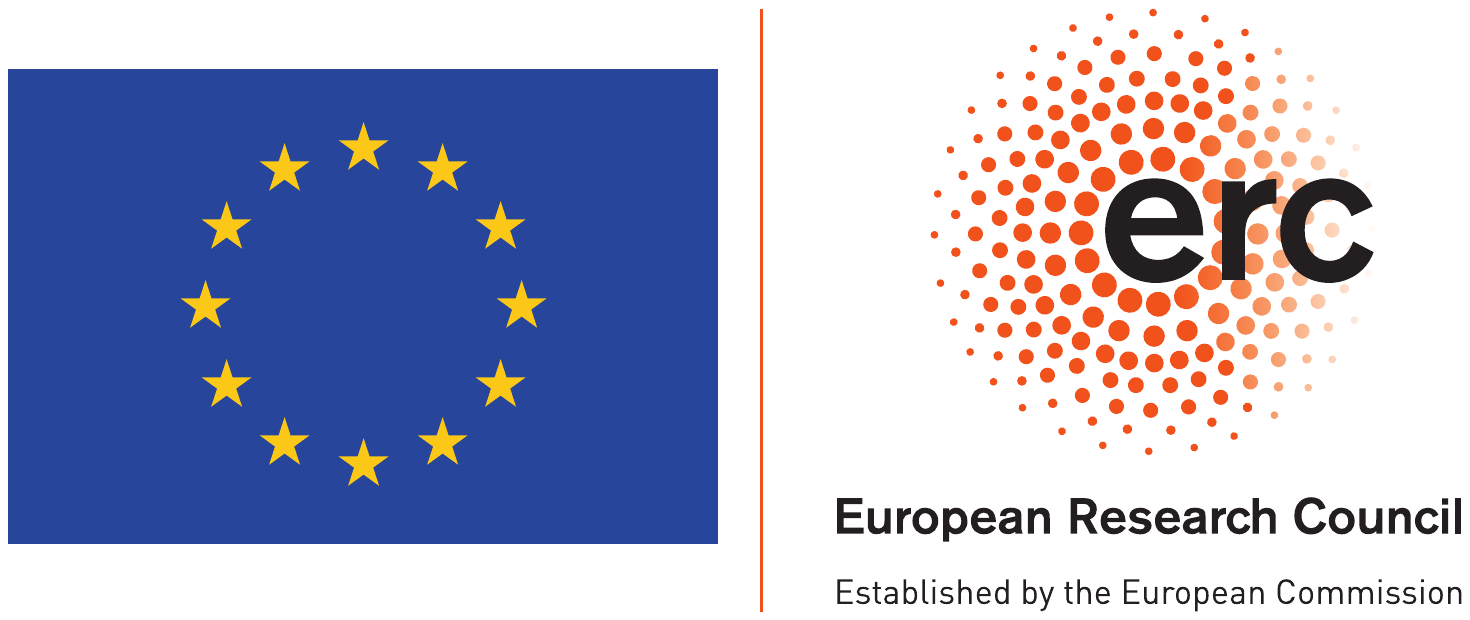}

\bibliography{strings,local,lib,proc}
\bibliographystyle{icml2024}

%%%%%%%%%%%%%%%%%%%%%%%%%%%%%%%%%%%%%%%%%%%%%%%%%%%%%%%%%%%%%%%%%%%%%%%%%%%%%%%
%%%%%%%%%%%%%%%%%%%%%%%%%%%%%%%%%%%%%%%%%%%%%%%%%%%%%%%%%%%%%%%%%%%%%%%%%%%%%%%
% APPENDIX
%%%%%%%%%%%%%%%%%%%%%%%%%%%%%%%%%%%%%%%%%%%%%%%%%%%%%%%%%%%%%%%%%%%%%%%%%%%%%%%
%%%%%%%%%%%%%%%%%%%%%%%%%%%%%%%%%%%%%%%%%%%%%%%%%%%%%%%%%%%%%%%%%%%%%%%%%%%%%%%
 \newpage
 \appendix
 %\onecolumn

\section{Human-centered AutoML and Machine Learning Operations (MLOps)} \label{app:mlops}
%MLOps focuses on the automation and monitoring of machine learning pipelines, whereas AutoML automates their configuration.
Both MLOps and AutoML aim to make machine learning more accessible and more convenient for potential users. Where AutoML aims at automating machine learning processes - especially the configuration of machine learning pipelines - MLOps is centered around the technical aspects as well as the deployment and monitoring of machine learning models in production.
In Figure~\ref{fig:crisp}, the tasks often associated with MLOps are represented in the right circle (though MLOps extends into experimentation and validation as well through e.g., reproducibility of experiments). By integrating AutoML within an MLOps framework, we can streamline the entire lifecycle of a machine learning project -- from data preparation to model deployment and monitoring -- eventually leading to \emph{AutoMLOps}. In view of Hypothesis 3, this strong automation nevertheless comes with a multitude of challenges in which humans are still needed, including defining business objectives (e.g., choosing success metrics) and problem framing, data management, interpreting results, making strategic decisions based on monitoring, ethical oversight, compliance, and continuous improvement by feedback loops. The multitude of tools coming together to form such a complex (semi-) automated system requires strong modularity of components, interfaces between components and users, compatibility between tools, and data management on experiments and versions; a good MLOps setup therefore becomes increasingly important.

\section{Selected AutoML Users with Different Needs} \label{app:users}
Overall, the users of AutoML are diverse and different applications demand different functionality of AutoML and specifically different types of user interactions. To put emphasis on this variety, in the following we showcase four distinct examples for potential users and their applications:
\begin{description}
    \item[A manufacturing expert in predictive maintenance:] Machine learning can reduce maintenance costs and unexpected downtime of heavy machinery through predictive maintenance. While this can easily be formulated as a classification, regression or survival analysis problem, domain expertise is important to create an effective machine learning model. Domain experts provide important information for feature selection and engineering, about constraints and even when it comes to evaluation of a machine learning model through their technical expertise and knowledge of the circumstances surrounding the machinery. While something like root cause analysis can be interesting in its own right, the predictive maintenance task usually only concerns itself with predictive performance and does not care about the interpretability of predictions.
    \item[A biologist working with gene or gene-expression data:] In such tasks, ML models are often adapted through domain knowledge. This domain knowledge usually comes in the form of string (or graph) kernels, gene (dis)similarity functions, or information from e.g., the gene ontology database (grouping of genes, and a hierarchical / DAG structure on groups). It is well known that using such domain knowledge can improve model performance~\cite{zhou-genebio19}. In terms of goals, users are, in many cases, interested in sparse solutions (i.e., few genes or few groups), as these results drive subsequent (expensive) experiments. %Interpretability sometimes is a goal, but more often it is not; so for a better juxtaposition to the next domain, let us assume it is not.
    Interpretability is often a goal, but not always.
    \item[A data scientist in a bank predicting credit risk:]
    Here, in contrast to the previous example, less customized ML models are often used, e.g., often linear models or other interpretable models; sometimes nonlinear forests or boosting-type models. Next to high predictive performance (maximizing profits), explainability of the machine learning model is often either a (legal) hard constraint or at least highly desirable. This needs to be incorporated into the AutoML process because otherwise very complex models, which are only marginally better than a simple one, might be selected. An additional consideration is fairness, which is notoriously hard to automate and quantify in single metrics~\cite{weerts-jair24}. 
    \item[A data scientist conducting unsupervised learning:] In contrast to supervised learning, it is notoriously hard to a-priori define ideal performance metrics that can simply be optimized against. Experienced data scientists usually diagnose and evaluate such models through visualization. In this case, an interactive approach along the lines of preference learning would be useful in AutoML systems for such tasks, see e.g.,~\cite {giovanelli-aaai24a}.
\end{description}

\section{Scalability and Cost of Human Efforts}\label{app:cost}

In the age of deep learning and large models, scalability is indeed a very important topic to consider in the context of AutoML. A lot of current AutoML research is centered around scalability to large models (incl. LLMs). On the one hand, AutoML and especially NAS algorithms are often designed to identify configurations with high-predictive power as well as energy or memory-efficient pipelines through, e.g., multi-objective optimization. On the other hand, research focuses on specific optimization methods like multi-fidelity optimization or scaling laws to make the search for a desirable pipeline as efficient as possible. We identify this as a current challenge for AutoML and call for further efficiency improvements to accommodate current architectures. \cite{tornede-arxiv23a} provides an in-depth discussion on this for LLMs.

Two main factors influence the final quality during model selection and configuration: Prior knowledge and available budget for optimization. The tighter the available budget, the more important the reliance on prior knowledge becomes -- one-shot configuration would be an extreme case. Unfortunately, many AutoML systems (with few modern exceptions) do not really allow the incorporation of very flexible and custom priors, which reflect human domain knowledge.
We can already see that in the domain of large (language) models. When, for example, configuring an LLM, there are only so many attempts that can realistically be made at finding a good architecture and/or hyperparameters, so – due to a lack of better alternatives – we often rely on human experts here. Even worse, the training of a large model over several days or weeks requires human supervision, e.g., to prevent divergence of training – in fact, no one would risk training a model on thousands of GPUs for a month without actively monitoring it. Automating this process as much as possible is still desirable. 

As a concrete example of a synergetic workflow, let us assume that only a single training of a large model is feasible and there is a constant risk of divergence. We envision a future in which a new kind of human-centered AutoML provides suggestions on how to adapt training when being at risk, but human supervision is necessary in view of the lack of extrapolation capabilities of AutoML for unseen future training steps. There are very first steps in this direction with approaches such as dynamic algorithm configuration~\citep{adriaensen-jair22a}.

Comparing the two extremes, manual vs. fully automated systems, it should be obvious that both lack the advantages of the other. AutoML will support users in tedious and error-prone tasks, such as manually trying out a sequence of different architectures or hyperparameters, whereas users will support AutoML in providing prior knowledge on high-level concepts and adapting overall objectives. Combining the complementary strengths of both can minimize the overall cost of designing and training a complex model. 

\section{Bias through Human Intervention}\label{app:bias}

The question of whether bias will be introduced through human intervention is an interesting but subtle topic. First of all, just because a decision process is based on data, ML and automation, this does not imply it is free of bias, in the sense of ``bad'' or ``improper'' human bias, as humans usually select the data source, humans specify goals, humans specify side constraints, and humans interpret results. 
We know that this can lead to drastic problems, if not properly accounted for, e.g., from the current fair-ML literature, but even earlier, there are many instances in applied data science where this has been observed. 
Automation can partially mitigate such biases, but it never fully eliminates them. And unfortunately, there is also a counter-effect of automation. If we accept that these biases can and will occur in many instances, and there is no silver bullet to avoid them a-priori, the next best thing is careful analysis, checking, and auditing. This is done by human experts, but the more automated, the less transparent a system is, the harder this becomes. In a fully autonomous system, a human expert is more detached from the machine learning process and may be able to fulfill oversight responsibilities to better prevent bias. However, with the increasing automation of black boxes and lack of transparency, it becomes harder for human experts to spot biases and related problems. Additionally, automation bias, the tendency to place too much trust in automated recommendations~\cite{skitka-ichcs99a}, may be reinforced through AutoML~\cite{weerts-jair24}. Finally, the previous point more or less refers to better, potential output of AutoML systems – to enable better human auditing. When it comes to humans selecting configurations during an interactive process, this can both introduce human biases or mitigate them. For example, a human could artificially drive the system to an unfair solution – or detect that the currently created solution is unfair and intervene. It is our belief that a transparent process is necessary to minimize this as much as possible. Nevertheless, this will not be a perfect system simply because humans are not perfect. However, removing human oversight and reason from the equation is, in our opinion, far more dangerous.


% You can have as much text here as you want. The main body must be at most $8$ pages long.
% For the final version, one more page can be added.
% If you want, you can use an appendix like this one.  

% The $\mathtt{\backslash onecolumn}$ command above can be kept in place if you prefer a one-column appendix, or can be removed if you prefer a two-column appendix.  Apart from this possible change, the style (font size, spacing, margins, page numbering, etc.) should be kept the same as the main body.
%%%%%%%%%%%%%%%%%%%%%%%%%%%%%%%%%%%%%%%%%%%%%%%%%%%%%%%%%%%%%%%%%%%%%%%%%%%%%%%
%%%%%%%%%%%%%%%%%%%%%%%%%%%%%%%%%%%%%%%%%%%%%%%%%%%%%%%%%%%%%%%%%%%%%%%%%%%%%%%

\end{document}